\documentclass[letterpaper, 10 pt, conference]{ieeeconf}  

\IEEEoverridecommandlockouts                              

\usepackage{amsmath, amsfonts, physics, bm, amsthm}

\newcommand{\fB}{\bm{f}}
\newcommand{\gB}{\bm{g}}

\newcommand{\sB}{\bm{s}}

\newcommand{\AB}{\bm{A}}

\newcommand{\CB}{\bm{C}}

\newcommand{\KB}{\bm{K}}

\newcommand{\MB}{\bm{M}}

\newcommand{\QB}{\bm{Q}}

\newcommand{\SB}{\bm{S}}

\newcommand{\VB}{\bm{V}}
\newcommand{\WB}{\bm{W}}
\newcommand{\XB}{\bm{X}}
\newcommand{\YB}{\bm{Y}}
\newcommand{\ZB}{\bm{Z}}

\newcommand{\calD}{\mathcal{D}}

\usepackage{multirow}
\let\labelindent\relax
\usepackage{enumitem}
\usepackage{xcolor}
\usepackage{graphicx}
\usepackage{booktabs}
\usepackage{makecell}
\usepackage{url}
\usepackage{xurl}

\def\mathbi#1{\textbf{\em #1}}

\title{\LARGE \bf
A Generalizable Physics-guided Causal Model for Trajectory Prediction in Autonomous Driving
}
\IEEEpeerreviewmaketitle

\author{Zhenyu Zong$^{1}$, Yuchen Wang$^{1}$, Haohong Lin$^{2}$, Lu Gan$^{3}$ and Huajie Shao$^{1}$
\thanks{$^{1}$Zhenyu Zong, Yuchen Wang and Huajie Shao are with Department of Computer Science, 
        William \& Mary, Williamsburg, VA 23185, USA
        {\tt\small \{zzong, ywang142, hshao\}@wm.edu}}%
\thanks{$^{2}$Haohong Lin is with SafeAI Lab, College of Engineering,
        Carnegie Mellon University, Pittsburgh, PA 15213, USA
        {\tt\small haohongl@andrew.cmu.edu}}%
\thanks{$^{3}$Lu Gan is with School of Aerospace Engineering,
        Georgia Institute of Technology, Atlanta, GA 30332, USA
        {\tt\small lgan@gatech.edu}}%
}

\begin{document}
\maketitle
\thispagestyle{empty}
\pagestyle{empty}
\begin{abstract}
Trajectory prediction for traffic agents is critical for safe autonomous driving. However, achieving effective zero-shot generalization in previously unseen domains remains a significant challenge. Motivated by the consistent nature of kinematics across diverse domains, we aim to incorporate domain-invariant knowledge to enhance zero-shot trajectory prediction capabilities. The key challenges include: 1) effectively extracting domain-invariant scene representations, and 2) integrating invariant features with kinematic models to enable generalized predictions. To address these challenges, we propose a novel generalizable Physics-guided Causal Model (PCM), which comprises two core components: a \textit{Disentangled Scene Encoder}, which adopts intervention-based disentanglement to extract domain-invariant features from scenes, and a \textit{CausalODE Decoder}, which employs a causal attention mechanism to effectively integrate kinematic models with meaningful contextual information. Extensive experiments on real-world autonomous driving datasets demonstrate our method's superior zero-shot generalization performance in unseen cities, significantly outperforming competitive baselines. The source code is released at \url{https://github.com/ZY-Zong/Physics-guided-Causal-Model}.
\end{abstract}    
\section{Introduction}



\label{sec:intro}
Trajectory prediction aims to forecast the future paths of dynamic agents in autonomous driving scenarios~\cite{afshar2024pbp, lange2024scene, huang2022multi, Generalization_Frenet}, which is critical for ensuring driving safety. Recent studies~\cite{dal2024joint,nayakanti2023wayformer} have primarily developed machine learning (ML) approaches for end-to-end trajectory prediction by leveraging multiple data modalities, including images, LiDAR, polylines, and waypoints. However, purely data-driven ML is prone to spurious correlation~\cite{CaDeT}, thus struggling to generalize to unseen domains.



To address this problem, existing works have developed various domain generalization techniques, such as transfer learning~\cite{yang2023rmp, cheng2023forecast} and trajectory tokenization for next-token prediction~\cite{NEURIPS2024smart}.
While these methods contribute to enhancing domain generalization, several challenges remain. Transfer learning approaches are often computationally expensive and struggle with zero-shot trajectory prediction. Additionally, trajectory tokenization techniques require massive data to effectively bridge the domain gap, making them less feasible in data-scarce or rapidly changing environments. Moreover, these purely data-driven ML approaches may not comply with physical laws. To deal with these issues, recent studies~\cite{gan2024goal, mao2023phy, geng2023physics} have developed physics-guided ML that incorporates system dynamics to enhance the robustness and generalizability of trajectory prediction. However, existing approaches often struggle to capture meaningful contextual information in dynamic and complex environments. In addition, none of them focuses on zero-shot prediction across two different domains.




\begin{figure}[tb]
    \centering
    \includegraphics[width=\linewidth]{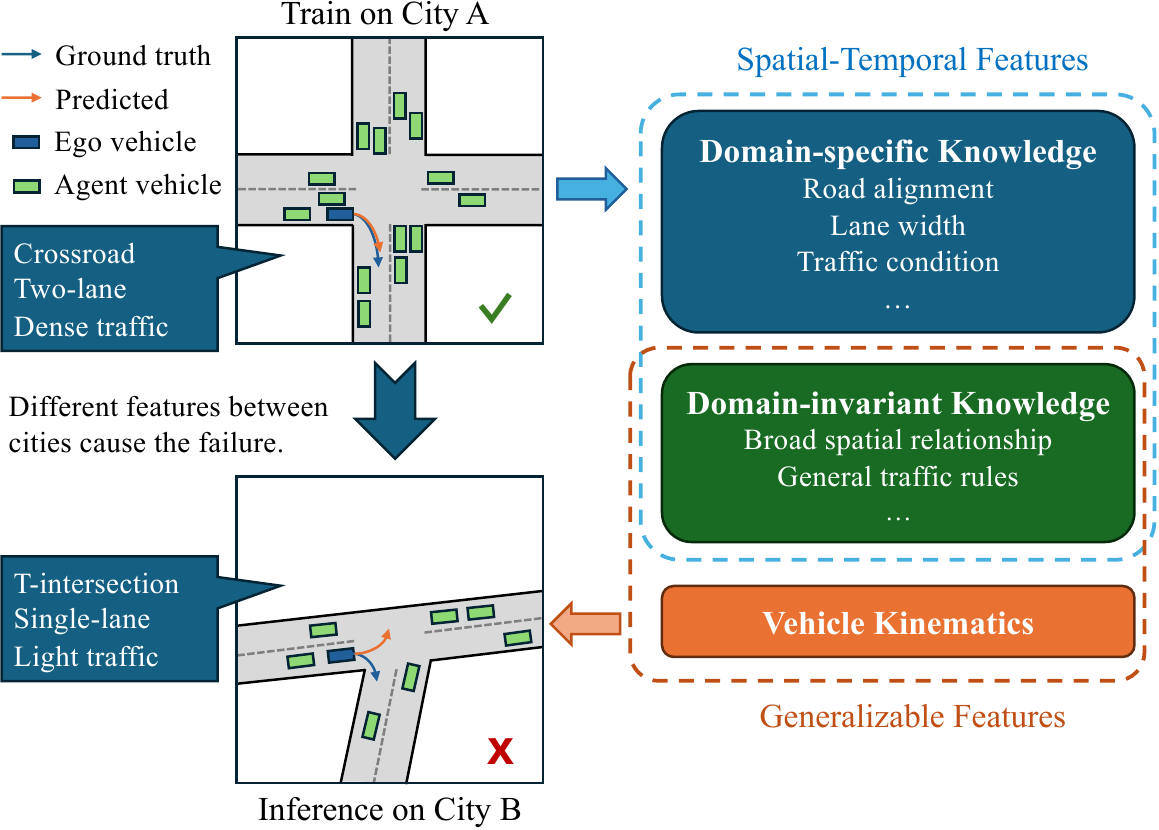}
    \vspace{-0.25in}
    \caption{Illustration of two right turn scenarios in two different cities. The model trained on city A can make correct prediction, but fail in unseen city B with different lane width, traffic condition and road alignment. To solve this problem, our method leverages two generalizable knowledge: \textbf{domain-invariant knowledge} separated from the spatial-temporal features and \textbf{vehicle kinematics}, to enhance domain generalization ability.}
    \label{fig:motivation}
    \vspace{-0.2in}
\end{figure}

To bridge this gap, we propose a generalizable Physics-guided Causal Model (PCM) that integrates domain-invariant features with vehicle kinematics to enhance zero-shot trajectory prediction in unseen domains. As shown in Fig.~\ref{fig:motivation}, models trained in City A often fail to generalize to scenarios in an unseen City B due to discrepancies in road alignments, lane widths, and traffic conditions. These characteristics represent domain-specific knowledge derived from spatial-temporal distributions. In contrast, broad spatial patterns such as road topology, and general traffic regulations are typically consistent across different urban environments. Motivated by this observation, we hypothesize that domain-invariant knowledge remains stable across cities, and that predicted trajectories should additionally conform to vehicle kinematic constraints to ensure physical plausibility and real-world deployability. The key challenges we are addressing include: (i) effectively extracting domain-invariant scene representations from map polylines, and (ii) integrating domain-invariant representations with kinematics to enhance the zero-shot generalizability of trajectory prediction.

To overcome the first challenge, we develop a \textit{Disentangled Scene Encoder} to learn domain-invariant features through intervention-based disentanglement. To address the second challenge, we introduce a \textit{CausalODE Decoder} that devises a causal attention mechanism to integrate invariant features and kinematics-guided predictions. Specifically, to comply with physical laws, we incorporate a two-wheel kinematic model~\cite{tumu2023physics} into neural ordinary differential equation (ODE)~\cite{chen2018neural} to learn vehicle dynamics and then use it to initialize a trajectory query for the decoder. Besides, inspired by causal representation learning~\cite{CaDeT,CaST}, we design a causal attention mechanism based on the Transformer decoder~\cite{vaswani2017attention} to fuse domain-invariant features and kinematics for capturing the interactions between them, thereby enhancing zero-shot generalization capability.




To evaluate the performance of our method, we implement two zero-shot experiments on real-world driving datasets: i) trained on nuPlan~\cite{nuplan} and evaluated on nuScenes~\cite{nuscenes}, and ii) trained on WOMD~\cite{waymo_2021_ICCV} and evaluated on nuPlan. Both of them show that our method significantly outperforms baselines in zero-shot trajectory predictions, suggesting its superior generalization capability.


\textbf{Our contributions} include: 1) we propose a novel generalizable Physics-guided Causal Model (PCM) that can enhance zero-shot trajectory prediction performance in unseen domains; 2) we introduce a Disentangled Scene Encoder to learn domain-invariant scene representations and a CausalODE Decoder to integrate learned invariant representations with kinematics to significantly enhance model's generalization capability; and 3) we present extensive experiments to validate the superior zero-shot prediction performance of our approach using real-world trajectory prediction datasets.


    


\section{Related Works}
\label{sec:related_work}

\textbf{Domain Generalization of Trajectory Prediction.}
Enhancing domain generalization of trajectory prediction remains an open research question. Existing techniques leverages transfer learning~\cite{ullrich2024transfer} 
together with random masks~\cite{yang2023rmp, cheng2023forecast} to tackle this challenge. However, these methods require high pretraining costs and finetuning models on new domains. In addition, they struggle with out-of-distribution (OOD) scenarios~\cite{meta_learning_challenges_2023, knowledge_distill_survey_2021}.
SMART~\cite{NEURIPS2024smart} leverages discrete representation to bridge different domains. However, it requires massive data to construct a generalizable token vocabulary. In addition, the discrepancy between the true value and its discretized representation can accumulate, which is harmful over long trajectories.
A most recent study~\cite{APE_li2024} has proposed Adaptive Prediction Ensemble (APE) that combines a rule-based expert and a learning-based expert to improve OOD generalization across datasets. However, since constant velocity rule-based expert struggle with predicting curved trajectories accurately, APE has to select from two inferior predictions when its learning-based expert cannot understand unseen turning scenarios. Different from existing works, we propose a new method that combines invariant scene features and kinematics to enhance the zero-shot generalization ability.

\textbf{Physics-guided ML for Trajectory Prediction.}
Recent studies~\cite{geng2023physics, tischmann2024physics, gan2024goal, mao2023phy} have explored physics-guided ML methods for generalized representation learning in trajectory prediction. For example, Geng et al.~\cite{geng2023physics} proposed a physics-informed hybrid model that combines a data-driven Transformer with the Intelligent Driver Model (IDM) to predict final trajectories. Tischmann et al.~\cite{tischmann2024physics} incorporated a physics-informed loss based on IDM to guide neural network training. Other works~\cite{gan2024goal, mao2023phy} embedded physical principles directly into machine learning architectures to enhance model generalization. While these methods demonstrate improved performance through the integration of physics, their evaluations are limited to within the same dataset. Moreover, they often fail to capture generalized contextual information from dynamic and complex environments, which may lead to poor predictions in unseen domains. To tackle this issue, we introduce a Disentangled Scene Encoder for extracting domain-invariant contextual features and a causal attention decoder for fusing these features with kinematics.

\textbf{Causal Representation Learning in Autonomous Driving.} Some research also employed causal representation learning based on Structural Causal Models (SCMs)~\cite{CaDeT, chen2021humantrajectorypredictioncounterfactual,CaST,Lin_2024} to improve generalization capability of trajectory prediction. For instance, CausalHTP~\cite{chen2021humantrajectorypredictioncounterfactual} utilized causal graphs and counterfactual interventions to avoid environment biases that mislead prediction results. CaDeT~\cite{CaDeT} and CaST~\cite{CaST} proposed disentanglement approaches to separate spurious parts from causal relations and then de-confound spurious causality. However, these methods either pre-define the causal structure or require causal annotations like~\cite{roelofs2023causalagents}.
To tackle these issues, a recent work, FUSION~\cite{Lin_2024}, leverages safety-aware causal Transformer to learn causality of Reinforcement Learning (RL) inputs. 
Liu et al.~\cite{rahimi2025sim} adopted sim-to-real causal transfer to learn multi-agent causal interactions without real-world causal annotations. CCDiff~\cite{lin2024causal} automatically identify and inject causal structures into the diffusion model. Unlike previous works, 
we introduce a causal intervention-based technique to disentangle domain-invariant features from scenes without needing causal annotations and pre-defined causal structures. These features are then integrated with vehicle kinematics through the attention block to learn causal interactions between context features and physics.

\section{Proposed Method}
\label{sec:method}

\subsection{Problem statement}
The objective is to predict the future trajectory of the ego agent in a multi-agent autonomous driving scenario consisting of \mbox{$n-1$} neighboring agents. Let \mbox{$\calD_{H} =\{\XB_1, \XB_2, \cdots, \XB_n\} $} represent the observed trajectories of these agents from time steps $t_0$ to $t_{H}$ in history, where $\XB_i$ denotes the observed trajectories of agent $i$, namely, $\XB_i=[\sB_i(t_0),\dots,\sB_i(t_H)]^\top$. For agent $i$, $\mathbi{s}_i(t_j)=[x_i(t_j),y_i(t_j),\theta_i(t_j),v_i(t_j)]^\top$ represents its state, including the position, heading, and velocity at time step $j$. Let $\MB$ denotes map polylines, which is represented by coordinate positions, directions, and lane types. Given environmental context $\MB$ and observed trajectories $\calD_H$, the goal of this work is to predict future trajectory of the ego agent $e$ in prediction horizon $T$, denoted by $\YB_e=[\sB_e(t_{H+1}),\dots,\sB_e(t_{H+T})]^\top$.

\subsection{Overall architecture}

\begin{figure*}[tb]
    \centering
    \includegraphics[width=0.8\linewidth]{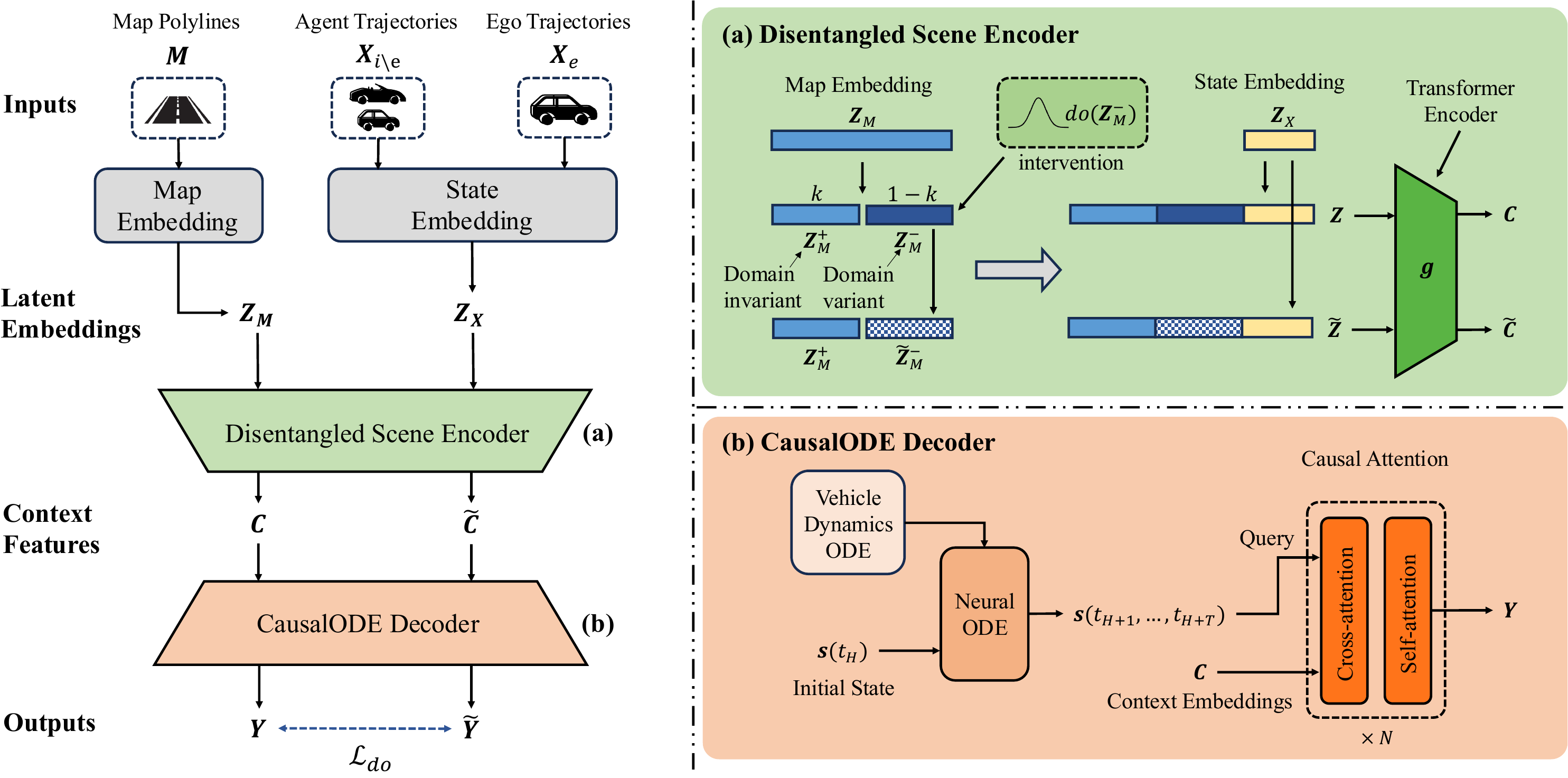}
    \vspace{-0.15in}
    \caption{Overall framework of the proposed method. It comprises two main parts: (a) a Disentangled Scene Encoder aiming to extract domain-invariant features; (b) a CausalODE decoder that integrates domain-invariant features with vehicle kinematics learned by neural ODE.}
    \label{fig:overview}
    \vspace{-0.2in}
\end{figure*}

To achieve domain generalization of the predictor, we develop a new physics-guided causal model to enhance zero-shot generalization capability, as shown in Fig.~\ref{fig:overview}. The core idea is to disentangle domain-invariant features from scenes and then integrate them with kinematics for trajectory prediction. The proposed method comprises two main components: (i) Disentangled Scene Encoder, aiming to learn domain-invariant representations from scenes; and (ii) CausalODE Decoder that captures the interactions between invariant features and vehicle dynamics.



\subsection{Disentangled scene encoder}
First, we adopt a disentangled scene encoder to learn domain-invariant features from map polylines, and then feed them into the scene encoder. By extracting invariant features, the scene encoder is able to generate meaningful context representations devoid of domain-specific information, e.g. different layout of unimportant map polylines in different cities. As a result, the trajectory decoder would rely exclusively on domain-invariant features to perform zero-shot prediction.


As shown in Fig.~\ref{fig:overview}, we take observed trajectories of agents $\XB$ and map polylines $\MB$ as inputs. Both inputs are respectively encoded into feature embeddings $\ZB_X$ and $\ZB_M$ using the multilayer perceptron (MLP). As illustrated in Fig.~\ref{fig:overview} (a), in order to learn invariant features from map polylines, we learn disentangled scene representation with the intervention mechanism that separates $\ZB_M$ into domain-invariant and domain-variant groups, denoted by $\ZB_M^+$ and $\ZB_M^-$ respectively. We set the first $k$ percent of latent features to be domain-invariant and the remaining to be domain-variant. Inspired by causal invariance in Structural Causal Model (SCM)~\cite{CaDeT}, disentangling can be achieved by cutting the causal relation between $\ZB_M^+$ and $\ZB_M^-$ with the intervention denoted by $do(\ZB_M^-)$. The core idea is to replace $\ZB_M^-$ with an intervention set to simulate revoking domain-variant features. In this work, assuming latent features follow multivariate Gaussian distribution, we construct an intervention set with data sampled from standard Gaussian distribution. 

After intervention, both original and intervened map features are concatenated with the vehicle trajectory representation $\ZB_X$ to obtain fused modality embeddings, respectively. Both embeddings are then fed into a transformer-based scene encoder $\gB_{\theta_1}$~\cite{jaegle2021perceiver} to obtain meaningful context information $\CB,\tilde{\CB}\in\mathbb{R}^{d\times h}$ respectively, where $d$ is the spatial latent dimension size and $h$ is the hidden latent dimension size.

\subsection{CausalODE decoder}

To further enhance the generalizability of trajectory prediction, we introduce a CausalODE Decoder that integrates domain-invariant features with kinematics. It consists of two parts: 1) a physics-guided neural ODE that serves as query based on vehicle dynamics, and 2) a causal attention module based on Transformer decoder~\cite{nayakanti2023wayformer} that captures interactions among context features and vehicle dynamics.

\textbf{Physics-guided neural ODE:} We aim to integrate kinematics into the neural ODE framework to learn the underlying governing equations. The main idea of neural ODE is to use deep neural networks (DNN) to approximate the vector field as $\dv{\mathbf{\sB}(t)}{t}=\Phi(\mathbf{\sB}(t),t,\phi),$
where $\Phi$ is the vector field parameterized by DNN and $\sB(t)$ is system states at time $t$. 

Recall that the vehicle state $\sB(t)$ contains four attributes, i.e., $(x, y, \theta, v)$, which represents the position of vehicle wheel center, heading, and velocity respectively. Following prior works~\cite{tumu2023physics,bevgpt}, we use the bicycle kinematic model to model vehicle dynamics as follows:
\begin{align}
    \label{eq:vehicle_dynmaics}
    \Phi(\mathbi{s}(t),t,\phi)=
    \begin{bmatrix}
    \dot{x} \\
    \dot{y} \\
    \dot{\theta}\\
    \dot{v}
    \end{bmatrix}
    =
    \begin{bmatrix}
        v\cos\theta\\
        v\sin\theta\\
        \frac{v\tan\delta}{L}\\
        a
    \end{bmatrix}.
\end{align}
$\delta$, $a$, $L$ represent steering angle, acceleration, and wheelbase length of the vehicle. These unknown dynamics terms corresponds to the third and fourth rows in Equation~(\ref{eq:vehicle_dynmaics}), and are learned via an MLP. Specifically, for each mode $i$, the terms $(\dot{\theta}_i, \dot{v}_i)$ are predicted by an MLP as $(\dot{\theta}_i, \dot{v}_i) = \text{MLP}_i(x_i, y_i, \theta_i, v_i)$. These learned components vary across vehicles and over time, capturing individual behavior and temporal dynamics.

Given the above vehicle dynamics, and let the trajectory’s initial state be the current observed state $\sB(t_H)$, trajectory prediction over a time horizon $T$ can be formulated as an initial value problem (IVP), and then solved by:
\begin{equation}\label{eq:ivp}
\begin{aligned}
\bm{s}(t)
&= \bm{s}(t_H)
  + \int_{t_H}^{t}
      \Phi\bigl(\bm{s}(\tau),\,\tau,\,\phi\bigr)\,\mathrm{d}\tau,\\
\bm{s}(t_{H+j})
&= \operatorname{ODESolve}\bigl( 
      \Phi,\;\bm{s}(t_H),\;t_H,t_{H+j}
  \bigr),\quad\\
  \text{for }j&=1,\dots,T.
\end{aligned}
\end{equation}


To capture uncertainty of unknown system coefficients in Eq.~\eqref{eq:vehicle_dynmaics}, we use $m$ neural ODEs with different parameters to learn vehicle dynamics in complex driving scenarios such as left turn and right turn. 


\textbf{Causal attention module:} Its inputs include outputs from neural ODEs, and context features generated by the scene encoder. To learn the causal relationships between them, we employ multi-head attention mechanism~\cite{nayakanti2023wayformer} to attend queries with context features. Specifically, for each prediction mode $j=1,\dots,m$ and spatial index $i=1,\dots,d$, the query, key and value for the decoder are: 
\begin{equation}
    \QB_j=\SB_j\WB_{\QB},\KB_i=\CB_i\WB_{\KB},\VB_i=\CB_i\WB_{\VB},
\end{equation}

where $\SB$ represents the mode prediction along horizon $T$. $\WB_{\QB}, \WB_{\KB}, \WB_{\VB}$ denotes the weight of query $\QB$, key $\KB$, and value $\VB$, respectively. If we set the hidden feature dimension of each mode prediction as $d_k$.
Then, the attention matrix $\AB$ among input nodes are: 
\begin{equation}
    \AB=\texttt{Softmax}(\QB\KB^\top/\sqrt{d_k}).
\end{equation}
Each value of attention matrix $\AB_{i,j}$ indicates the impact of context feature $\CB_i$ to each trajectory $\SB_j$. We then compute each head's output by weighting the values, concatenating the vectors, and projecting back yields the physics query output. We follow each cross‐attention block with a self‐attention block to enhance the physics query internal relationship. After repeating this process for $N$ layers, we we apply a final linear projection to produce the predicted trajectory $\YB$.




\subsection{Objective function}
The overall objective is composed of two loss functions: (i) intervention loss and (ii) Gaussian Mixture Model (GMM) loss commonly used in trajectory prediction~\cite{nayakanti2023wayformer, chai2019multipath, shi2022mtr}. We detail these two loss functions below.

\textbf{Intervention loss:} Recall that we feed both original and domain-invariant latent representations $\ZB$ and $\tilde{\ZB}$ through the encoder $\gB_{\theta_1}$, resulting in context features $\CB=\gB_{\theta_1}(\ZB)$ and $\tilde{\CB}=\bm{g}_{\theta_1}(\tilde{\ZB})$.
These two features are fed into the decoder, denoted by $\fB_{\theta_2}$, to predict trajectories $\YB$ and $\tilde{\YB}$, respectively. To learn domain-invariant features, we force these two predicted trajectories to be close to each other. Mathematically, we have the following objective function:
\begin{equation}
\label{eq:intervention loss}
    \mathop{\arg\min}\limits_{\theta_1,\theta_2} \mathcal{L}_{do}(\YB,\tilde{\YB})=\frac{1}{T}\sum_{j=H+1}^{H+T}\norm{\fB_{\theta_2}(\CB) - \fB_{\theta_2}(\tilde{\CB})}^2,
\end{equation}
where $H$ is the current time step and $T$ is the prediction horizon. $\gB_{\theta_1}$ and  $\fB_{\theta_2}$ denote the encoder and decoder, respectively.


\textbf{GMM loss:} During model training, we aim to find the nearest predicted trajectory compared with ground truth $\hat{\YB}$ and minimize their difference. Given $m$ different modes of predictions, define $\hat{i}$ as the nearest index. This give the best predtion $\YB_{\hat{i}}$ with probability $p_{\hat{i}}$ Then the GMM loss~\cite{chai2019multipath} contains classification and regression part as below:
\begin{equation}
\mathcal{L}_{\mathrm{GMM}}(\YB, \hat{\YB})
= \mathbb{E}_{(x,y)}\bigl[
\mathcal{L}_{\mathrm{reg}}\bigl(\YB_{\hat{i}},\hat{\YB}\bigr)
  + \mathcal{L}_{\mathrm{cls}}\bigl(p_{\hat{i}}\bigr)
\bigr],
\end{equation}
where $\mathcal{L}_{\mathrm{reg}}$ denote negative log-likelihood of the ground-truth trajectory under the bivariate Gaussian distribution of the selected nearest mode, and $\mathcal{L}_{\mathrm{cls}}$ denote the the cross entropy to maximize the probability of the selected trajectory~$\YB_{\hat{i}}$.



\textbf{Overall objective:} The final objective is the sum of the intervention loss and two Gaussian Mixture Model (GMM) loss:
\begin{equation}
    \mathcal{L}=\mathcal{L}_{GMM}(\YB, \hat{\YB})+\mathcal{L}_{GMM}(\tilde{\YB}, \hat{\YB})+\lambda\mathcal{L}_{do}(\YB,\tilde{\YB}),
\end{equation}
where $\lambda$ is a hyperparameter that balances the third term.

\section{Experiments}
\label{sec:experiment}

\subsection{Datasets}
We evaluate our method using public dataset described below. In data preprocessing, we use ScenarioNet~\cite{scenarionet_2023} and unitraj~\cite{zhu2024unitraj} to generated unified descriptions of three different datasets to achieve cross-dataset evaluation.
Specifically, (1) \textbf{nuScenes} dataset~\cite{nuscenes} contains 850 train-validation scenarios and 150 test scenarios collected from Boston and Singapore. (2) \textbf{nuPlan} dataset~\cite{nuplan} provides a broader coverage, with data from Boston, Singapore, Las Vegas, and Pittsburgh. Since nuScenes and nuPlan share city maps, we use only the Pittsburgh subset (35,058 scenes) for training in nuPlan to prevent overlap during cross-dataset evaluation, while the original nuPlan test split (20,756 scenes from all four cities) is used for testing. 
(3) \textbf{Waymo Open Motion Dataset (WOMD)}~\cite{waymo_2021_ICCV} offers a significantly larger scale, with 487K training scenes and 44K test scenes collected from six cities including San Francisco, Phoenix, Mountain View, Los Angeles, Detroit, and Seattle—all of which are unseen in nuPlan. To reduce computational costs, we randomly sample 10\% of the training scenes from WOMD for training.



\subsection{Implementation details}
For feature embeddings, we use 1-layer and 2-layer MLP respectively to project agents trajectories and map polylines into hidden spaces with size of 240. We select the closest 384 map polylines around the ego agent, with up to 20 points for each polyline. For disentanglement, we set domain-invariant percentage to $k=0.5$. We will also explore the effect of $k$ on model performance in Section~\ref{sec:ablations}. The weight of intervention loss is $\lambda=1$.

For scene encoder, we use a Perceiver-like~\cite{jaegle2021perceiver} architecture with 1 cross-attention layer followed by 2 self-attention layers. For the decoder, we use 6 3-layer MLPs with hidden size of 200 to parameterize ODEs and solve IVP with Dopri5 solver~\cite{lienen2022torchode}. The causal attention block consists of 8 causal attention layers.

For all experiments, models are trained 100 epochs with AdamW optimizer and OneCycleLR learning rate scheduler on 4 Nvidia RTX A5000 GPUs.

\subsection{Metrics}
We use four unified metrics in unitraj~\cite{zhu2024unitraj} for evaluation. For all metrics, we consider the top $k=6$ prediction modes. Specifically, we use the following metrics:

\begin{itemize}[leftmargin=*]

\item \textbf{Minimum Average / Final Displacement Error (minADE / minFDE)}: It computes the average / final time step of L2 distance error between ground truth and the closest trajectory among 6 modes of predictions:

\item \textbf{Miss Rate (MR)}: The prediction is defined to miss when final trajectory displacement exceeding 2 meters. Miss rate is the ratio of miss samples at prediction horizon $T$.

\item \textbf{Brier minimum Final Displacement Error (brier-minFDE)}: It accounts for probability of the best predicted trajectory $p$ in minFDE by adding the penalty value $(1-p)^2$ to the L2 distance.
\end{itemize}

\subsection{Baselines}
To assess the performance of our method, we not only compare it against competitive trajectory prediction baselines, but also some advanced methods with generalization capabilities. Specifically,
\textbf{Wayformer}~\cite{nayakanti2023wayformer} fuses multi-modal input features with self-attention encoder to generate spatial-temporal relationship, which is attended to predict trajectories through cross-attention decoder.
\textbf{Autobot}~\cite{girgis2021autobot} uses multi-head self-attention modules to learn social and temporal interactions, allowing scene-consistent multi-agent trajectory predictions.
\textbf{MTR}~\cite{shi2022mtr} integrates global intention localization and local movement refinement with a unified transformer, which ensures the model to learn agent interactions and predict scene-compliant predictions.
\textbf{G2LTraj}~\cite{zhang2024g2ltraj} simultaneously generates local intermediate trajectory predictions between global key steps with spatial-temporal constraints.
\textbf{APE}~\cite{APE_li2024} selects either learning-based or rule-based prediction expert with the highest ranking score for trajectory prediction.
\textbf{Forecast-MAE}~\cite{cheng2023forecast} randomly masks both agent trajectories and map elements, which are used to pretrain a scene-level encoder.
\textbf{RMP}~\cite{yang2023rmp} randomly masks historical agent trajectories and pretrains an agent-centric motion encoder that learns motion dynamics and temporal continuity.
\textbf{SMART}~\cite{NEURIPS2024smart} models agent trajectories and vectorized map as discrete sequence tokens for next-token prediction via a decoder-only transformer.

\begin{figure*}[!tb]
    \centering
    \includegraphics[width=0.8\linewidth]{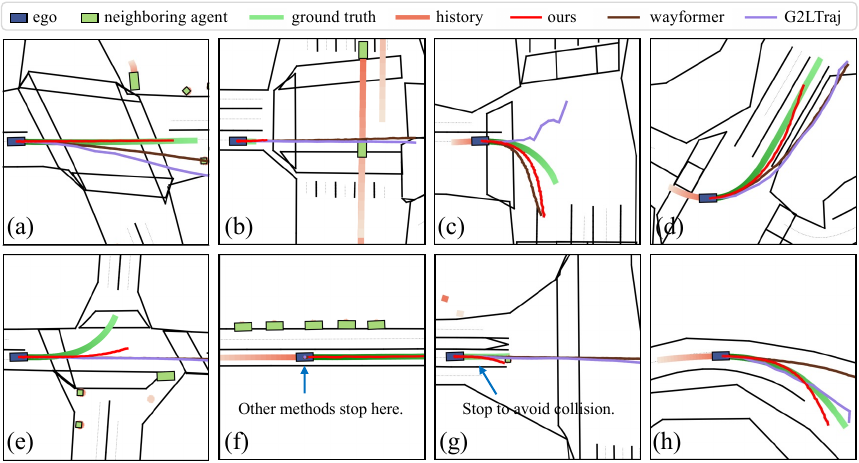}
    \caption{The trajectory prediction visualization for top-three methods on \textbf{8} different nuScenes scenarios. (a) Our method drives normally, while others crash into pedestrians. (b) Our method stops before the pavement in the crossroad with heavy traffic, while two other predictions crash into other vehicles. (c) Our method turns right smoothly, while G2LTraj turns left. G2LTraj and Wayformer's predictions cannot follow vehicle kinematics well in this scenario. (d) Our method turns left without driving out of the road. (e) Our method correctly turns left at the crossroad. (f) Our method drives normally with vehicles parking aside the road, while other methods stop. (g) Our method stops to avoid collision to the pedestrian ahead. (h) Our method turns right smoothly, indicating it follows vehicle kinematics.}
    \label{fig:case_visualization}
    \vspace{-0.2in}
\end{figure*}

\begin{table}[!tb]
  \centering
  \caption{\textbf{nuPlan-to-nuScenes:} Performance comparison between our method and baselines for zero-shot evaluation on nuScenes, trained on nuPlan dataset. Results are averaged over three random seeds. \textbf{Bold} represents the best results. }
  \label{tab:nuplan2nuscenes}
  \begin{tabular}{l c c c c}
    \toprule
    Method                 & $\mathrm{minADE}_6$ & $\mathrm{minFDE}_6$ & \makecell{brier\\-minFDE} & \makecell{Miss\\Rate} \\
    \midrule
    Wayformer~\cite{nayakanti2023wayformer} & 1.036  & 2.621  & 3.299  & 0.469 \\
    AutoBot~\cite{girgis2021autobot}        & 1.197  & 2.782  & 3.430  & 0.465 \\
    G2LTraj~\cite{zhang2024g2ltraj}         & 1.188  & 2.754  & 3.419  & 0.467 \\
    MTR~\cite{shi2022mtr}                   & 1.360  & 3.236  & 3.877  & 0.600 \\\midrule
    Forecast-MAE~\cite{cheng2023forecast}   & 1.159  & 2.673  & 3.974  & 0.602 \\
    RMP~\cite{yang2023rmp}                  & 1.485  & 3.460  & 4.032  & 0.653 \\
    SMART~\cite{NEURIPS2024smart}           & 1.812  & 3.587  & 4.513  & 0.695 \\
    APE~\cite{APE_li2024}                   & 3.655  & 9.214  & 9.214  & 0.889 \\
    \midrule
    Ours                   & \textbf{0.897} & \textbf{2.205} & \textbf{2.857} & \textbf{0.398}  \\
    \bottomrule
  \end{tabular}
\end{table}

\begin{table}[!tb]
  \centering
  \caption{\textbf{WOMD-to-nuPlan:} Performance comparison between our method and baselines for zero-shot evaluation on nuplan, trained on WOMD dataset. Results are averaged over three random seeds. \textbf{Bold} represents the best results.}
  \label{tab:womd2nuplan}
  \begin{tabular}{l c c c c}
    \toprule
    Method                 & $\mathrm{minADE}_6$ & $\mathrm{minFDE}_6$ & \makecell{brier\\-minFDE} & \makecell{Miss\\Rate} \\
    \midrule
    Wayformer~\cite{nayakanti2023wayformer}    & 0.745  & 2.034  & 2.635  & 0.317 \\
    AutoBot~\cite{girgis2021autobot}           & 0.851  & 2.308  & 2.934  & 0.340 \\
    G2LTraj~\cite{zhang2024g2ltraj}            & 0.872  & 2.324  & 2.938  & 0.388 \\
    MTR~\cite{shi2022mtr}                      & 0.791  & 2.125  & 2.621  & 0.394 \\\midrule
    Forecast-MAE~\cite{cheng2023forecast}      & 0.936  & 2.531  & 2.985  & 0.407 \\
    RMP~\cite{yang2023rmp}                     & 1.375  & 2.832  & 3.247  & 0.419 \\
    SMART~\cite{NEURIPS2024smart}              & 1.614  & 4.082  & 4.626  & 0.492 \\
    APE~\cite{APE_li2024}                      & 3.122  & 8.373  & 8.373  & 0.766 \\
    \midrule
    Ours     & \textbf{0.720} & \textbf{1.951} & \textbf{2.562} & \textbf{0.315}  \\
    \bottomrule
  \end{tabular}
  \vspace{-0.2in}
\end{table}

\subsection{Comparison against the state-of-the-art}
We compare the zero-shot generalization performance of our method with baselines. Specifically, we train models on one dataset with 21 history time steps and zero-shot evaluate them on another with 60 prediction time steps. 
The performance are evaluated on the top \mbox{$k=6$} prediction modes with four unified metrics \textbf{minADE}, \textbf{minFDE}, \textbf{brier-minFDE} and \textbf{miss rate} in unitraj~\cite{zhu2024unitraj}.

\textbf{Evaluation on nuScenes and trained on nuPlan}.
As shown in Table~\ref{tab:nuplan2nuscenes}, our method outperforms the second best baseline by a margin of 0.139, 0.416, 0.442, and 0.067 in terms of minADE, minFDE, brier-minFDE and miss rate respectively. We attribute the improvement to physics knowledge and domain-invariant features. It reveals that our method can learn these two generalized knowledge so that the model maintains the performance well in unseen scenarios. The state-of-art generalizable trajectory prediction method, APE, fails to achieve good performance in our experiments, indicating it may not fully understand the road conditions in unseen domains. As a result, its routing function struggles to select the best prediction expert for trajectory prediction. In addition, our method outperforms Forecast-MAE, RMP and SMART, indicating our method's superior zero-shot generalization ability.

\textbf{Evaluation on nuPlan and trained on WOMD}.
Compared to nuScenes and nuPlan, WOMD is an even larger-scale dataset which contains more complicated driving data by covering more turning and fewer straight scenarios. As shown in Table~\ref{tab:womd2nuplan}, our method still achieves the best performance in complex scenarios with smaller minADE, minFDE, brier-minFDE and miss rate values of 0.025, 0.083, 0.059 and 0.002 respectively. It is noticeable that baselines like wayformer and MTR, are able to achieve promising trajectory prediction performance in unseen datasets. This indicates that, with sufficient training data, the spatial-temporal relationships learned by their models may contain generalizable information. Nevertheless, our method improves it further by learning vehicle dynamics and domain-invariant features. Forecast-MAE and RMP fail to deliver strong performance because randomly masking out data within a single dataset does not guarantee the learning of generalizable features that can effectively transfer to other datasets. This risks overfitting to dataset-specific patterns while ignoring transferable motion semantics. Similarly, SMART underperforms due to its tokenization strategy, which may lead to the loss of fine-grained motion nuances critical for precise trajectory forecasting.

\subsection{Ablation study}
\label{sec:ablations}
\textbf{Impact of key components}. We also investigate the impact of two important components: (a) disentangled scene encoder with intervention loss, and (b) CausalODE decoder, on trajectory prediction performance using nuScenes dataset, as shown in Table~\ref{tab:block_ablations}. We can observe that the performance drops 0.157, 0.514, 0.543 and 0.1 for minADE, minFDE, berierFDE and miss rate, respectively, when removing physics-guided ODE. This reveals learning physical dynamics can boost the generalization. Similarly, removing disentanglement and intervention loss from our method gives worse performance, which shows the effectiveness of disentanglement to generate domain-invariant features. Finally, the result without both components shows that fusion of physical dynamics and domain-invariants features can attend their relations well and boost model generalization capability.

\textbf{Impact of domain-invariant percentage}.
Recall that we control the percentage of domain-invariant features in the disentangled scene encoder as illustrated in Table~\ref{tab:block_ablations} with the hyperparameter k. We investigate the impact of k on the nuScenes dataset to assess its impact on model performance. As shown in Table VII, the performance degrades when k is set either too small or too large. When k is too small, the scene encoder fails to learn latent representations since most map embeddings are revoked. Conversely, when k is too large, the domain-variant components lack sufficient latent dimensions and are captured within the domain-invariant feature groups. In our experiments, we choose k = 0.5, which gives the best performance.

\begin{figure}[!tb]
    \centering
    \includegraphics[width=0.8\linewidth]{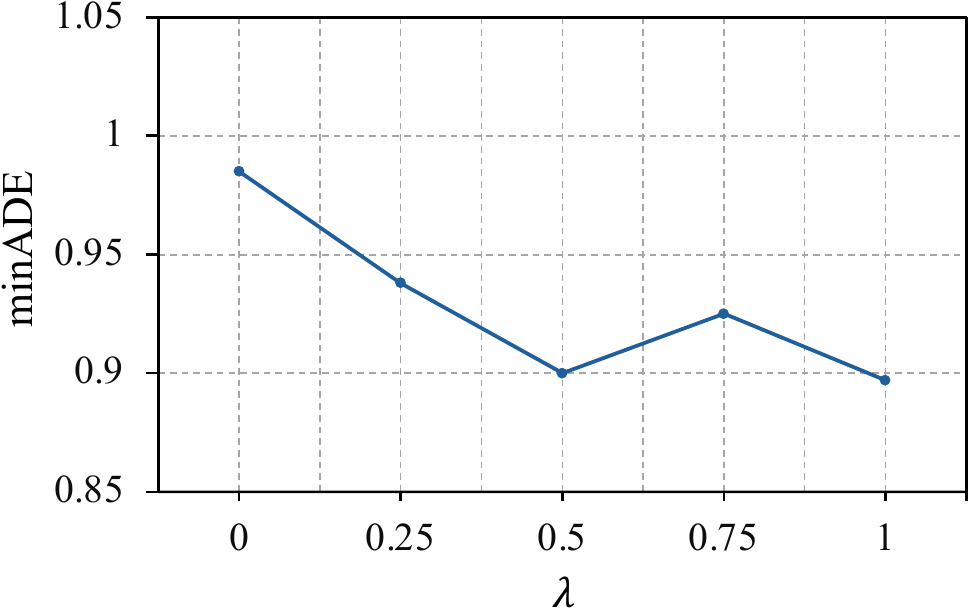}
    \vspace{-0.1in}
    \caption{Parameter sensitivity of $\lambda$.}
    \label{fig:sensitivity}
\end{figure}

\textbf{Parameter sensitivity}. We conduct experiments on the nuPlan-to-nuScenes setting to analyze the sensitivity of parameter $\lambda$ with values sampled from $\{0, 0.25, 0.5, 0.75, 1\}$. As shown in Fig.~\ref{fig:sensitivity}, minADE remains relatively stable when $\lambda >0$, but increases significantly when $\lambda=0$. Overall, the proposed method get the best performance with $\lambda=1$, so we set $\lambda=1$ in our experiments.

\begin{table}[!tb]
\centering
\caption{Ablation studies of the proposed method. Experiments are trained on nuPlan and evaluated on nuScenes.}
\vspace{0.02in}
\label{tab:block_ablations}
\begin{tabular}{ll|llll}
\toprule
(a) & (b) & $\mathrm{minADE}_6 $ & $\mathrm{minFED}_6$ & brier-minFDE & Miss Rate \\\midrule
$\times$ & $\times$ & 1.484   & 3.907   & 4.567    & 0.667 \\
$\times$ & $\checkmark$ & 1.031   & 2.652   & 3.309    & 0.400 \\
$\checkmark$ & $\times$ & 1.010   & 2.526   & 3.196    & 0.453 \\\midrule
\multicolumn{2}{c|}{Ours}       & \textbf{0.853}   & \textbf{2.012}   & \textbf{2.653}    & \textbf{0.353}     \\
\bottomrule
\end{tabular}
\end{table}

\begin{table}[!tb]
\centering
\caption{Impact of domain-invariant percentage $k$. Experiments are trained on nuPlan and evaluated on nuScenes.}
\label{tab:k_sensitivity}
\begin{tabular}{l|llll}
\toprule
$k$ & $\mathrm{minADE}_6$ & $\mathrm{minFED}_6$ & brier-minFDE & Miss Rate \\\midrule
0.2                           & 1.115   & 2.756   & 3.381    & 0.533     \\
0.4                           & 0.999   & 2.573   & 3.252    & 0.480     \\
0.6                           & 0.972   & 2.439   & 3.105    & 0.493     \\
0.8                           & 1.012   & 2.560   & 3.228    & 0.487     \\\midrule
0.5 (Ours)                    & \textbf{0.853}   & \textbf{2.012}   & \textbf{2.653}    & \textbf{0.353}     \\
\bottomrule
\end{tabular}
\vspace{-0.2in}
\end{table}

\subsection{Qualitative examples}
In Fig.~\ref{fig:case_visualization}, we visualize the best predicted trajectories of top three methods in eight unseen scenarios in nuScenes. In case (a), where the traffic condition at the intersection is relatively clear, our method executes a normal driving maneuver, whereas Wayformer and G2LTraj fail to detect pedestrians and collide with them. In case (b), cross-street traffic is actively moving. Our model is able to stop safely before the pedestrian crossing, while the baselines overlook the traffic context and crash into surrounding vehicles. In case (c), both Wayformer and our method correctly execute a right turn consistent with the ground truth, while G2LTraj erroneously turns left. Notably, our predicted trajectory is smoother than those from the baselines, highlighting the effectiveness of the vehicle kinematic constraints embedded in CausalODE. Similarly, the smooth right turn depicted in case (h) further supports our model’s compliance with physical vehicle dynamics. Case (d) illustrates our model's ability to safely execute a left turn without veering off the road. In case (e), our method correctly turns left to avoid both pedestrians on the right and a vehicle ahead, suggesting a strong understanding of spatial interactions in the scene. In case (f), despite parked vehicles along the roadside, our method proceeds normally, while baseline models halt unnecessarily. Finally, in case (g), our model appropriately stops to avoid a pedestrian directly ahead, demonstrating context-aware behavior.

In summary, these cases reveal that our method is generalizable to unseen scenarios by learning domain-invariant knowledge and vehicle kinematics. 

\subsection{Inference time}
To guarantee safety in autonomous driving, the inference speed of the model should satisfy real-time requirements. In Table~\ref{tab:inference_time}, we investigate the inference speed of our method. Though vehicle kinematics increase the processing time, our method takes 37.23 ms, which still satisfy real-time requirement~\cite{luo2019time} within 100 ms.

\begin{table}[!tb]
  \centering
  \caption{Inference latency comparison between our method and baselines. We measure all the model's one mode prediction time on one RTX A5000 GPU (batch size=1), averaged over the whole nuScense test dataset.}
  \begin{tabular}{lccccc}
    \toprule
    Method & Wayformer & AutoBot & G2LTraj & MTR  & Ours  \\
    \midrule
    Latency (ms) & 13.93 & \textbf{10.05} & 10.16 & 65.00 & 37.23 \\
    \midrule
    Method & F-MAE & RMP & SMART & APE \\
    \midrule
    Latency (ms) & 21.82 & 12.46 & 11.75 & 42.38 \\
    \bottomrule
  \end{tabular}
  \vspace{-0.2in}
  \label{tab:inference_time}
\end{table}

\section{Conclusion}
\label{sec:conclusion}
In this work, we proposed a novel generalizable physics-guided causal model to enable zero-shot trajectory prediction performance by learning domain-invariant features and physical dynamics. Our method is composed of two core components: the disentangled scene encoder to effectively extract domain-invariant parts of scene representations, and the CausalODE decoder to fuse domain-invariant features with physics-guided trajectory queries. Extensive experimental results on real-world driving datasets demonstrated the superior zero-shot generalizability of our method over baselines.

Despite these promising results, our current implementation relies on a simplified bicycle model, which may not fully capture the complex dynamics of real-world vehicles. In future work, we plan to incorporate more sophisticated kinematic models to further enhance the physical fidelity.

\section*{Acknowledgments}
Research reported in this paper was sponsored in part by NSF CPS 2311086, NSF CIRC 716152, NSF RITEL 2506890, NAIRR 250288, and Faculty Research Grant at William \& Mary 141446.



\bibliographystyle{IEEEtran}
\bibliography{IEEEabrv, references}  

@INPROCEEDINGS{nuscenes,
  title={nuScenes: A multimodal dataset for autonomous driving},
  author={Holger Caesar and Varun Bankiti and Alex H. Lang and Sourabh Vora and 
          Venice Erin Liong and Qiang Xu and Anush Krishnan and Yu Pan and 
          Giancarlo Baldan and Oscar Beijbom}, 
  booktitle={CVPR},
  year=2020
}

@INPROCEEDINGS{nuplan, 
  title={NuPlan: A closed-loop ML-based planning benchmark for autonomous vehicles},
  author={H. Caesar, J. Kabzan, K. Tan et al.},
  booktitle={CVPR ADP3 workshop},
  year=2021
}

@InProceedings{waymo_2021_ICCV,
  author={Ettinger, Scott and Cheng, Shuyang and Caine, Benjamin and Liu, Chenxi and Zhao, Hang and Pradhan, Sabeek and Chai, Yuning and Sapp, Ben and Qi, Charles R. and Zhou, Yin and Yang, Zoey and Chouard, Aur'elien and Sun, Pei and Ngiam, Jiquan and Vasudevan, Vijay and McCauley, Alexander and Shlens, Jonathon and Anguelov, Dragomir},
  title={Large Scale Interactive Motion Forecasting for Autonomous Driving: The Waymo Open Motion Dataset},
  booktitle= {Proceedings of the IEEE/CVF International Conference on Computer Vision (ICCV)},
  month={October},
  year={2021},
  pages={9710-9719}
}

@inproceedings{afshar2024pbp,
  title={Pbp: Path-based trajectory prediction for autonomous driving},
  author={Afshar, Sepideh and Deo, Nachiket and Bhagat, Akshay and Chakraborty, Titas and Shao, Yunming and Buddharaju, Balarama Raju and Deshpande, Adwait and Motional, Henggang Cui},
  booktitle={2024 IEEE International Conference on Robotics and Automation (ICRA)},
  pages={12927--12934},
  year={2024},
  organization={IEEE}
}

@inproceedings{lange2024scene,
  title={Scene informer: Anchor-based occlusion inference and trajectory prediction in partially observable environments},
  author={Lange, Bernard and Li, Jiachen and Kochenderfer, Mykel J},
  booktitle={2024 IEEE International Conference on Robotics and Automation (ICRA)},
  pages={14138--14145},
  year={2024},
  organization={IEEE}
}

@inproceedings{huang2022multi,
  title={Multi-modal motion prediction with transformer-based neural network for autonomous driving},
  author={Huang, Zhiyu and Mo, Xiaoyu and Lv, Chen},
  booktitle={2022 International Conference on Robotics and Automation (ICRA)},
  pages={2605--2611},
  year={2022},
  organization={IEEE}
}

@inproceedings{Generalization_Frenet,
   title={Improving the Generalizability of Trajectory Prediction Models with Frenét-Based Domain Normalization},
   DOI={10.1109/icra48891.2023.10160788},
   booktitle={2023 IEEE International Conference on Robotics and Automation (ICRA)},
   publisher={IEEE},
   author={Ye, Luyao and Zhou, Zikang and Wang, Jianping},
   year={2023},
   month=may, pages={11562–11568}
}

@inproceedings{ullrich2024transfer,
author={Ullrich, Lars and McMaster, Alex and Graichen, Knut},
booktitle={2024 35th IEEE Intelligent Vehicles Symposium (IV)},
title={Transfer Learning Study of Motion Transformer-based Trajectory Predictions},
address={Jeju Island, Korea},
year={2024},
pages={110--117},
doi={10.1109/IV55156.2024.10588422},
publisher={IEEE}
}

@misc{APE_li2024,
      title={Adaptive Prediction Ensemble: Improving Out-of-Distribution Generalization of Motion Forecasting}, 
      author={Jinning Li and Jiachen Li and Sangjae Bae and David Isele},
      year={2024},
      eprint={2407.09475},
      archivePrefix={arXiv},
      primaryClass={cs.RO},
      url={https://arxiv.org/abs/2407.09475}, 
}

@misc{meta_learning_challenges_2023,
      title={Advances and Challenges in Meta-Learning: A Technical Review}, 
      author={Anna Vettoruzzo and Mohamed-Rafik Bouguelia and Joaquin Vanschoren and Thorsteinn Rögnvaldsson and KC Santosh},
      year={2023},
      eprint={2307.04722},
      archivePrefix={arXiv},
      primaryClass={cs.LG},
      url={https://arxiv.org/abs/2307.04722}, 
}

@InProceedings{knowledge_distill_survey_2021,
author="Nguyen, Dang
and Gupta, Sunil
and Nguyen, Trong
and Rana, Santu
and Nguyen, Phuoc
and Tran, Truyen
and Le, Ky
and Ryan, Shannon
and Venkatesh, Svetha",
editor="Oliver, Nuria
and P{\'e}rez-Cruz, Fernando
and Kramer, Stefan
and Read, Jesse
and Lozano, Jose A.",
title="Knowledge Distillation with Distribution Mismatch",
booktitle="Machine Learning and Knowledge Discovery in Databases. Research Track",
year="2021",
publisher="Springer International Publishing",
address="Cham",
pages="250--265"
}

@INPROCEEDINGS{CaDeT,
  author={Pourkeshavarz, Mozhgan and Zhang, Junrui and Rasouli, Amir},
  booktitle={2024 IEEE/CVF Conference on Computer Vision and Pattern Recognition (CVPR)}, 
  title={CaDeT: A Causal Disentanglement Approach for Robust Trajectory Prediction in Autonomous Driving}, 
  year={2024},
  volume={},
  number={},
  pages={14874-14884},
  doi={10.1109/CVPR52733.2024.01409}}

@misc{chen2021humantrajectorypredictioncounterfactual,
      title={Human Trajectory Prediction via Counterfactual Analysis}, 
      author={Guangyi Chen and Junlong Li and Jiwen Lu and Jie Zhou},
      year={2021},
      eprint={2107.14202},
      archivePrefix={arXiv},
      primaryClass={cs.CV},
      url={https://arxiv.org/abs/2107.14202}, 
}

@misc{CaST,
      title={Deciphering Spatio-Temporal Graph Forecasting: A Causal Lens and Treatment}, 
      author={Yutong Xia and Yuxuan Liang and Haomin Wen and Xu Liu and Kun Wang and Zhengyang Zhou and Roger Zimmermann},
      year={2023},
      eprint={2309.13378},
      archivePrefix={arXiv},
      primaryClass={cs.LG},
      url={https://arxiv.org/abs/2309.13378}, 
}

@article{Lin_2024,
   title={Safety-Aware Causal Representation for Trustworthy Offline Reinforcement Learning in Autonomous Driving},
   volume={9},
   ISSN={2377-3774},
   url={http://dx.doi.org/10.1109/LRA.2024.3379805},
   DOI={10.1109/lra.2024.3379805},
   number={5},
   journal={IEEE Robotics and Automation Letters},
   publisher={Institute of Electrical and Electronics Engineers (IEEE)},
   author={Lin, Haohong and Ding, Wenhao and Liu, Zuxin and Niu, Yaru and Zhu, Jiacheng and Niu, Yuming and Zhao, Ding},
   year={2024},
   month=may, pages={4639–4646} }

@inproceedings{scenarionet_2023,
author = {Li, Quanyi and Peng, Zhenghao and Feng, Lan and Liu, Zhizheng and Duan, Chenda and Mo, Wenjie and Zhou, Bolei},
title = {ScenarioNet: open-source platform for large-scale traffic scenario simulation and modeling},
year = {2023},
publisher = {Curran Associates Inc.},
address = {Red Hook, NY, USA},
booktitle = {Proceedings of the 37th International Conference on Neural Information Processing Systems},
articleno = {172},
numpages = {27},
location = {New Orleans, LA, USA},
series = {NIPS '23}
}

@inproceedings{nayakanti2023wayformer,
  title={Wayformer: Motion forecasting via simple \& efficient attention networks},
  author={Nayakanti, Nigamaa and Al-Rfou, Rami and Zhou, Aurick and Goel, Kratarth and Refaat, Khaled S and Sapp, Benjamin},
  booktitle={2023 IEEE International Conference on Robotics and Automation (ICRA)},
  pages={2980--2987},
  year={2023},
  organization={IEEE}
}

@inproceedings{jaegle2021perceiver,
  title={Perceiver: General perception with iterative attention},
  author={Jaegle, Andrew and Gimeno, Felix and Brock, Andy and Vinyals, Oriol and Zisserman, Andrew and Carreira, Joao},
  booktitle={International conference on machine learning},
  pages={4651--4664},
  year={2021},
  organization={PMLR}
}

@article{chen2018neural,
  title={Neural ordinary differential equations},
  author={Chen, Ricky TQ and Rubanova, Yulia and Bettencourt, Jesse and Duvenaud, David K},
  journal={Advances in neural information processing systems},
  volume={31},
  year={2018}
}

@article{chai2019multipath,
  title={Multipath: Multiple probabilistic anchor trajectory hypotheses for behavior prediction},
  author={Chai, Yuning and Sapp, Benjamin and Bansal, Mayank and Anguelov, Dragomir},
  journal={arXiv preprint arXiv:1910.05449},
  year={2019}
}

@article{shi2022mtr,
  title={Motion transformer with global intention localization and local movement refinement},
  author={Shi, Shaoshuai and Jiang, Li and Dai, Dengxin and Schiele, Bernt},
  journal={Advances in Neural Information Processing Systems},
  volume={35},
  pages={6531--6543},
  year={2022}
}

@article{zhu2024unitraj,
  title={UniTraj: Universal human trajectory modeling from billion-scale worldwide traces},
  author={Zhu, Yuanshao and Yu, James Jianqiao and Zhao, Xiangyu and Wei, Xuetao and Liang, Yuxuan},
  journal={arXiv preprint arXiv:2411.03859},
  year={2024}
}

@article{girgis2021autobot,
  title={Latent variable sequential set transformers for joint multi-agent motion prediction},
  author={Girgis, Roger and Golemo, Florian and Codevilla, Felipe and Weiss, Martin and D'Souza, Jim Aldon and Kahou, Samira Ebrahimi and Heide, Felix and Pal, Christopher},
  journal={arXiv preprint arXiv:2104.00563},
  year={2021}
}

@article{zhang2024g2ltraj,
  title={G2LTraj: A global-to-local generation approach for trajectory prediction},
  author={Zhang, Zhanwei and Hua, Zishuo and Chen, Minghao and Lu, Wei and Lin, Binbin and Cai, Deng and Wang, Wenxiao},
  journal={arXiv preprint arXiv:2404.19330},
  year={2024}
}

@inproceedings{lienen2022torchode,
  title = {torchode: A Parallel {ODE} Solver for PyTorch},
  author = {Marten Lienen and Stephan G{\"u}nnemann},
  booktitle = {The Symbiosis of Deep Learning and Differential Equations II, NeurIPS},
  year = {2022},
  url = {https://openreview.net/forum?id=uiKVKTiUYB0}
}

@misc{
roelofs2023causalagents,
title={CausalAgents: A Robustness Benchmark for Motion Forecasting Using Causal Relationships},
author={Rebecca Roelofs and Liting Sun and Benjamin Caine and Khaled S. Refaat and Benjamin Sapp and Scott Ettinger and Wei Chai},
year={2023},
url={https://openreview.net/forum?id=9WdB5yVICCA}
}

@inproceedings{tumu2023physics,
  title={Physics constrained motion prediction with uncertainty quantification},
  author={Tumu, Renukanandan and Lindemann, Lars and Nghiem, Truong and Mangharam, Rahul},
  booktitle={2023 IEEE Intelligent Vehicles Symposium (IV)},
  pages={1--8},
  year={2023},
  organization={IEEE}
}

@inproceedings{yang2023rmp,
  title={Rmp: A random mask pretrain framework for motion prediction},
  author={Yang, Yi and Zhang, Qingwen and Gilles, Thomas and Batool, Nazre and Folkesson, John},
  booktitle={2023 IEEE 26th International Conference on Intelligent Transportation Systems (ITSC)},
  pages={3717--3723},
  year={2023},
  organization={IEEE}
}

@inproceedings{cheng2023forecast,
  title={Forecast-mae: Self-supervised pre-training for motion forecasting with masked autoencoders},
  author={Cheng, Jie and Mei, Xiaodong and Liu, Ming},
  booktitle={Proceedings of the IEEE/CVF International Conference on Computer Vision},
  pages={8679--8689},
  year={2023}
}

@article{gan2024goal,
  title={Goal-based Neural Physics Vehicle Trajectory Prediction Model},
  author={Gan, Rui and Shi, Haotian and Li, Pei and Wu, Keshu and An, Bocheng and Li, Linheng and Ma, Junyi and Ma, Chengyuan and Ran, Bin},
  journal={arXiv preprint arXiv:2409.15182},
  year={2024}
}

@article{mao2023phy,
  title={Phy-Taylor: Partially Physics-Knowledge-Enhanced Deep Neural Networks via NN Editing},
  author={Mao, Yanbing and Gu, Yuliang and Sha, Lui and Shao, Huajie and Wang, Qixin and Abdelzaher, Tarek},
  journal={IEEE Transactions on Neural Networks and Learning Systems},
  year={2023},
  publisher={IEEE}
}

@article{geng2023physics,
  title={A physics-informed transformer model for vehicle trajectory prediction on highways},
  author={Geng, Maosi and Li, Junyi and Xia, Yingji and Chen, Xiqun Michael},
  journal={Transportation research part C: emerging technologies},
  volume={154},
  pages={104272},
  year={2023},
  publisher={Elsevier}
}

@inproceedings{rahimi2025sim,
  title={Sim-to-Real Causal Transfer: A Metric Learning Approach to Causally-Aware Interaction Representations},
  author={Rahimi, Ahmad and Luan, Po-Chien and Liu, Yuejiang and Rajic, Frano and Alahi, Alexandre},
  booktitle={Proceedings of the IEEE/CVF Conference on Computer Vision and Pattern Recognition 2025 [forthcoming publication]},
  year={2025}
}

@article{lin2024causal,
  title={Causal Composition Diffusion Model for Closed-loop Traffic Generation},
  author={Lin, Haohong and Huang, Xin and Phan-Minh, Tung and Hayden, David S and Zhang, Huan and Zhao, Ding and Srinivasa, Siddhartha and Wolff, Eric M and Chen, Hongge},
  journal={arXiv preprint arXiv:2412.17920},
  year={2024}
}

@inproceedings{tischmann2024physics,
  title={Physics Informed Deep Learning for Motion Prediction in Autonomous Driving},
  author={Tischmann, Patrick and Baumann, Robin and Novo, Anne Stockem},
  booktitle={AmEC 2024--Automotive meets Electronics \& Control; 14. GMM Symposium},
  pages={7--12},
  year={2024},
  organization={VDE}
}

@article{dal2024joint,
  title={Joint Perception and Prediction for Autonomous Driving: A Survey},
  author={Dal'Col, Lucas and Oliveira, Miguel and Santos, V{\'\i}tor},
  journal={arXiv preprint arXiv:2412.14088},
  year={2024}
}

@article{vaswani2017attention,
  title={Attention is all you need},
  author={Vaswani, Ashish and Shazeer, Noam and Parmar, Niki and Uszkoreit, Jakob and Jones, Llion and Gomez, Aidan N and Kaiser, {\L}ukasz and Polosukhin, Illia},
  journal={Advances in neural information processing systems},
  volume={30},
  year={2017}
}

@ARTICLE{bevgpt,
  author={Wang, Pengqin and Zhu, Meixin and Zheng, Xinhu and Lu, Hongliang and Zhong, Hui and Chen, Xianda and Shen, Shaojie and Wang, Xuesong and Wang, Yinhai and Wang, Fei-Yue},
  journal={IEEE Transactions on Intelligent Vehicles}, 
  title={BEVGPT: Generative Pre-trained Foundation Model for Autonomous Driving Prediction, Decision-Making, and Planning}, 
  year={2024},
  volume={},
  number={},
  pages={1-13},
  doi={10.1109/TIV.2024.3449278}}

@article{luo2019time,
  title={Time constraints and fault tolerance in autonomous driving systems},
  author={Luo, Yujia},
  journal={Tech. rep, Tech. Rep},
  year={2019}
}

@inproceedings{NEURIPS2024smart,
 author = {Wu, Wei and Feng, Xiaoxin and Gao, Ziyan and Kan, Yuheng},
 booktitle = {Advances in Neural Information Processing Systems},
 editor = {A. Globerson and L. Mackey and D. Belgrave and A. Fan and U. Paquet and J. Tomczak and C. Zhang},
 pages = {114048--114071},
 publisher = {Curran Associates, Inc.},
 title = {SMART: Scalable Multi-agent Real-time Motion Generation via Next-token Prediction},
 volume = {37},
 year = {2024}
}


\end{document}